\documentclass[11pt,a4paper]{article}
\usepackage{graphicx}
\usepackage[]{emnlp2021}
\usepackage[american]{babel}
\usepackage[T1]{fontenc}
\usepackage{times}
\usepackage{url}
\usepackage{microtype}

\usepackage{mathtools}

\usepackage[latin1]{inputenc}
\usepackage[T1]{fontenc}
\usepackage{type1cm}
\usepackage[american]{babel}
\usepackage{url}
\usepackage{xspace}
\DeclareMathAlphabet{\mathcalbf}{OMS}{pzc}{b}{n}
\usepackage{microtype}

\newcommand{\Ni}{(1)~}
\newcommand{\Nii}{(2)~}
\newcommand{\Niii}{(3)~}

\newcommand{\Na}{(a)~}
\newcommand{\Nb}{(b)~}
\newcommand{\Nc}{(c)~}

\RequirePackage{color}

\DeclareGraphicsExtensions{.pdf,.ai,.jpg,.png}
\setkeys{Gin}{pagebox=artbox}

\graphicspath{{emnlp21-reframing-figures/}}

\newcommand{\bsfigure}[3][scale=1.0]{%
  \begin{figure}[tb]
    \centering
    \includegraphics[#1]{#2}
    \caption{#3}\label{#2}
  \end{figure}}

\usepackage{booktabs}
\newcommand{\strategy}{\ensuremath{\mathcal{S}}\xspace}

\RequirePackage{type1cm}
\RequirePackage{color}
\RequirePackage{soul}
\setstcolor{blue}

\newsavebox\bscombox
\newcommand{\bscom}[3][]{%
  \sbox{\bscombox}{\fontsize{8}{9}\selectfont#1#2#3}
  \noindent
  \st{#2}{\selectfont
    \color{blue}#3\ifx\\#1\\\else{\fontsize{8}{9}\selectfont\color{violet}[#1]}\fi
    }}

\begin{document}

\title{Controlled Neural Sentence-Level Reframing of News Articles}

\author{
Wei-Fan Chen \\
Paderborn University \\
Department of Computer Science \\
{\tt cwf@mail.upb.de} \\\And
Khalid Al-Khatib \\
Bauhaus-Universit\"at Weimar \\
Faculty of Media, Webis Group \\
{\tt khalid.alkhatib@uni-weimar.de } \\\AND
Benno Stein\\
Bauhaus-Universit\"at Weimar \\
Faculty of Media, Webis Group \\
{\tt benno.stein@uni-weimar.de} \\\And
Henning Wachsmuth \\
Paderborn University \\
Department of Computer Science \\
{\tt henningw@upb.de}
}

\date{}

\maketitle

\begin{abstract}
Framing a news article means to portray the reported event from a specific perspective, e.g., from an economic or a health perspective. {\em Reframing} means to change this perspective. Depending on the audience or the submessage, reframing can become necessary to achieve the desired effect on the readers. Reframing is related to adapting style and sentiment, which can be tackled with neural text generation techniques. However, it is more challenging since changing a frame requires rewriting entire sentences rather than single phrases. In this paper, we study how to computationally reframe sentences in news articles while maintaining their coherence to the context. We treat reframing as a sentence-level fill-in-the-blank task for which we train neural models on an existing media frame corpus. To guide the training, we propose three strategies: framed-language pretraining, named-entity preservation, and adversarial learning. We evaluate respective models automatically and manually for topic consistency, coherence, and successful reframing. Our results indicate that generating properly-framed text works well but with tradeoffs.
\end{abstract}

\section{Introduction}
\label{sec:introduction}

Framing is a rhetorical means to emphasize a perspective of an issue \cite{de:2005,chong:2007}. It is basically driven by argument selection \cite{ajjour:2019b} and, hence, it belongs to the inventio canon in particular \cite{aristotle:2004}. The media employ framing to reorient how audiences think \cite{chong:2007}, or to promote a decided interpretation. For example, when talking about a certain law one may emphasize its {\em economic} impact or its consequences regarding {\em crime}.

\begin{table}[t!]

\small
\centering
\setlength{\tabcolsep}{2pt}

\begin{tabular}{p{0.98\linewidth}}
\toprule
{\bfseries (a)~Economic Frame} (original text) \\
\midrule
Key Congressional backers of the measure, sponsored by Senator Alan K. Simpson, Republican of Arizona, and Romano L. Mazzoli, Democrat of Kentucky, wanted a flexible spending limit. 
\textbf{Implicit in the debate and the stalemate that left the bill to die when Congress adjourned was a recognition that the cost of immigration reform would be high, although no one knew how high.}
Without reform, though, the presence of what may be six million illegal aliens in this country exacts an economic and social toll. \\
\addlinespace
\midrule
{\bfseries (b)~Legality Frame} (reframed text) \\
\midrule
Key Congressional backers of the measure, sponsored by Senator Alan K. Simpson, Republican of Arizona, and Romano L. Mazzoli, Democrat of Kentucky, wanted a flexible spending limit. 
\textbf{``It's time for Congress to take action,'' says a spokesman for the bill's sponsors, who want a flexible spending limit.}
Without reform, though, the presence of what may be six million illegal aliens in this country exacts an economic and social toll. \\
\addlinespace
\midrule
{\bfseries (c)~Crime Frame} (reframed text) \\
\midrule
Key Congressional backers of the measure, sponsored by Senator Alan K. Simpson, Republican of Arizona, and Romano L. Mazzoli, Democrat of Kentucky, wanted a flexible spending limit. 
\textbf{``Illegal aliens' is a growing problem in the country,'' says a spokesman for the measure's sponsors.}
Without reform, though, the presence of what may be six million illegal aliens in this country exacts an economic and social toll. \\
\bottomrule
\end{tabular}

\caption{
(a)~Sample text from the media frames corpus \cite{card:2015}. The bold sentence is labeled with the {\em economic} frame.
Having reframed the sentence with the approach of this paper, the text remains largely coherent and topic-consistent while showing the {\em legality} frame~(b) and {\em crime} frame~(c), respectively.}
\label{table-reframing-example1}

\end{table}

Reframing means to change the perspective of an issue. It can be a strategy to communicate with opposing camps of audiences, and, sometimes, just replacing specific terms can be enough to reach a reframing effect. Consider in this regard a reporter who may prefer to use ``undocumented worker'' instead of ``illegal aliens'' in left-leaning news \cite{albert:2020}. While still referring to the same people, the former can provoke a discussion of the economic impact of hiring them; the latter may raise issues of crime and possible deportation. Such low-level style reframing has been studied in recent work \cite{chakrabarty:2021}.

Usually, reframing requires rewriting entire sentences rather than single words or phrases. Table~\ref{table-reframing-example1} illustrates the change of a sentence from the economic frame~(a) to the legality frame~(b) and the crime frame~(c). While the original text emphasizes the cost of immigration reform, the legality-framed text quotes that ``It's time for Congress to take action,'' and the crime-framed text includes the notion of ``illegal aliens''.%
\footnote{Ethical concerns regarding the correctness of reframed texts will be discussed in Section~\ref{subsection:limitations}.}
The terms ``bill'' and ``measure'' in the respective reframed versions ensure the topical coherence of the texts. Two facts become clear from the example, namely that reframing needs
\Ni
notable rewriting to shift the focus, and
\Nii
overlapped entities to ensure topic consistency. 

To work in the real world, a computational reframing model needs to be able to rewrite sentences completely. At the same time, the model has to preserve the context, by maintaining coherence and topic consistency. Towards these goals, we propose to treat reframing as a {\em sentence-level fill-in-the-blank} task: Given three consecutive sentences plus a target frame, mask the middle sentence and generate a sentence that connects the preceding and the succeeding sentence in a natural way and that conveys the target frame. This task implies three research questions: 
\Ni
How to tackle a sentence-level fill-in-the-blank task in general?
\Nii
How to generate a sentence with a specific frame?
\Niii
How to make the sequence of sentences coherent?

Sentence-level blank filling is a new and unsolved task. We approach this task via controlled text generation, that is, by tweaking input and output of a sequence-to-sequence model where the masked sentence is the target output, and the preceding and the succeeding sentences are the inputs (Section~\ref{sec:approach}). For the second and third research question, we propose three training strategies: 
\Na
{\em framed-language pretraining}, to finetune the model on all framed texts in order to learn the framed ``language'',
\Nb
{\em named-entity preservation}, to support the model in maintaining important entities extracted from the masked sentence, and
\Nc
{\em adversarial learning}, to show the model undesired output texts in order to learn to avoid them.

Based on the corpus of \newcite{card:2015} with annotated sentence-level frames (Section~\ref{sec:data}), we empirically evaluate the pros and cons of each strategy and of combinations thereof (Section~\ref{sec:experiments}). The results reveal that our approach changes sentences properly from the original to the target frame in most cases (Section~\ref{sec:discussion}). Some ``reframing directions'' remain challenging, such as from crime to economic. We find that obtaining high scores for all assessed dimensions at the same time is hard to achieve; for example, the adversarial learning strategy gives a strong signal towards the target frame at the expense of lower coherence. The implied trade-offs suggest that reframing technology should be configurable when applying it in a real-world scenario to put different stress on each sentence.

The contribution of this paper is threefold:
\Ni
We demonstrate that sentence-level reframing can be tackled as a fill-in-the-blank task.
\Nii
We propose three training strategies for controlled text generation problems such as reframing.
\Niii
We provide empirical insights into unresolved aspects of the computational reframing of news articles.

\section{Related Work}
\label{sec:section-sf}

Framing in media, particularly in news articles, has been investigated widely in the communication and journalism areas~\cite{entman:1993,de:2005,chong:2007}. It has been defined in different ways, ranging from a narrow view such as ``make moral judgments''~\cite {entman:1993} to a broader one including the ``interpretative packages''~\cite{gamson:1989}. The set of frames for a certain topic can be issue-specific or generic. For example, the possible issue-specific frames for the topic of {\em Internet} may include {\em online communication} and {\em online services}, whereas the generic ones include {\em economically optimistic} and {\em political criticism}~\cite{rossler:2001}. In this paper, we adopt the following narrow definition of frames: ``a frame is an emphasis in the salience of different aspects of a topic'' \cite{de:2005}.

In the area of natural language processing, media frame analysis is a relatively new topic. Most existing works adopt the frame definition in social science, where framing refers to \emph{a choice of perspective}~\cite{hartmann:2019}. A more specific definition, which targets the argumentation contexts, defines a frame as a \emph{set of arguments that share an aspect}~\cite{ajjour:2019b}. As for frame  classification, most of the proposed approaches~\cite{naderi:2017,hartmann:2019,khanehzar:2021} employ the media frames corpus~\cite{card:2015}, which is built upon the framing scheme of \citet{boydstun:2013}. Following these approaches, we utilize the media frames corpus to build the dataset for our task.
The study of media frames is closely related to the analysis of bias and unfairness conveyed by the media~\cite{chen:2020a,chen:2020b}. For example, \citet{chen:2018} observed that the (potentially frame-specific) word choice may directly make a news article appear to have politically left or right bias.

The only existing reframing approach that we are aware of is the one of \citet{chakrabarty:2021}. In that work, a new model for reframing is developed by identifying phrases indicative for specific frames, and then replacing phrases that belong to the source frame with some that belong to the target one. As such, most of the content of the reframed text is kept, and only a few words are replaced. In contrast, we deal with reframing at the sentence level, and we do not require parallel training pairs or a dictionary to correlate words and frames.

In principle, reframing can be seen as a style transfer task~\cite{shardlow:2014,shen:2017,chen:2018}. Research on text style transfer focus on the areas of sentiment transfer (e.g., replacing `gross' by `awesome')~\cite{shen:2017} and  text simplification (e.g., replacing `perched' by `sat')~\cite{shardlow:2014}. We applied recent style transfer models to our task~\cite{mai:2020,shen:2020}, observing that these models perform very poorly (e.g., generating unreadable text).

\section{Approach}
\label{sec:approach}

We now present our approach to sentence-level reframing. We discuss how we tackle the reframing problem as a fill-in-the-blank task, and we propose three training strategies to generate a sentence that is framed as desired and that fits to the surrounding text. Figure~\ref{approach} illustrates our approach.

\subsection{Reframing as a Fill-in-the-Blank Task}

As discussed in Section~\ref{sec:introduction}, reframing implies two problems: \Ni To rewrite entire sentences from a text as much as needed in order to encode a given target frame; and \Nii to maintain coherence and topic consistency with respect to the context given in the text. To tackle both problems simultaneously, we propose to treat reframing as a specific type of sentence-level fill-in-the-blank task. 

In particular, let a sequence of three contiguous sentences, $\langle s_1, s_2, s_3 \rangle$, be given along with a target frame, $f$. The middle sentence, $s_2$, is the sentence to be reframed, and the other two sentences define the context taken into account for $s_2$. The fill-in-the-blank idea is to mask $s_2$, such that we have $\langle s_1, ${\tt[MASK]}, $s_3 \rangle$. The task, in turn, is to then decode the masked token {\tt[MASK]} to $\hat{s}_{2,f}$, a variation of the sentence~$s_2$ that is reframed to~$f$~and both coherent and topic-consistent to~$s_1$ and~$s_3$. 

No proper solution exists for this task yet, and only little prior work has addressed closely related problems (see Section~\ref{sec:section-sf}). To approach the task, we propose a sequence-to-sequence model $r(\cdot)$ where the input is the two context sentences, $\langle s_1, s_3 \rangle$, and the output to be generated is $s_2$. In order to consider frame information in rewriting, we train one individual frame-specific model $r_f(\cdot)$ for each frame~$f$ from a given set of target frames, $F$, such that
\begin{equation}
\forall f \in F: r_{f}\left ( s_1, s_3 \right ) \sim \hat{s}_{2,f} 
\end{equation}

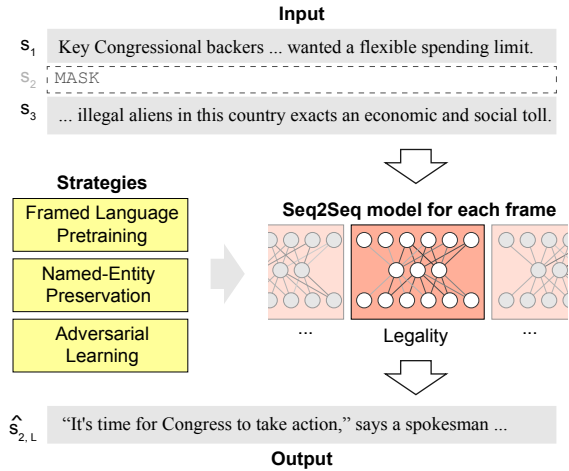
\bsfigure{approach}{Illustration of our approach. The sequence-to-sequence model trained on desired target frame (here, {\em Legality}) takes the context sentences ($s_1$, $s_3$) as input and $s_2$ as target output. After applying the three training strategies, the model learns to decode ${\tt[MASK]}$ to the text addressing ``It's time for Congress to take action''.}

\subsection{Training Strategies}
\label{subsection:method_training_strategies}

To better control the text generated by the model, we further guide the training process, by additionally considering the following three complementary training strategies. All three aim at providing extra information to the reframing model. In Section~\ref{sec:experiments}, we experiment with variations of the models to test each strategy and their combinations thoroughly.

\paragraph{Framed-Language Pretraining ($\strategy_{\rm F}$)} 

Due to the complexity of manual annotation, we can expect only a limited number of task instances for each frame $f \in F$ in practice, so the models may have insufficient knowledge about how to generate framed language. To mitigate this problem, the first strategy we propose is to {\em pretrain the reframing model on all available text of any frame $f \in F$}. After that, this pretrained model will be further fine-tuned using instances from one particular frame.

\paragraph{Named-Entity Preservation ($\strategy_{\rm N}$)} 

Given that a complete sentence is to be generated, a reframing model may mistakenly generate off-topic and incoherent text, if not controlled for. To avoid this, the second strategy is {\em to encode knowledge about the named entities to be discussed}. In particular, the set of named entities, $N$, can be extracted from $s_2$ and added to the input of the model.%
\footnote{We use the pretrained model {\em en\_core\_web\_lg} from spaCy for named entity recognition in our experiments.} 
Then, the input of the model can be extended to $s_1$ {\tt[NE]} $N$ {\tt[/NE]} $s_3$, where {\tt[NE]} and {\tt[/NE]} are special tokens to indicate the start and ending of named entities.

\paragraph{Adversarial Learning ($\strategy_{\rm A}$)} 

During training, the instances fed to the default model are all ``positive'' samples where the output $s_2$ comes from the same sentences $\langle s_1, s_2, s_3 \rangle$ the input sentences $s_1$ and~$s_3$ are from. While this helps learning to generate coherent text, it impedes learning reframing. For example, if the goal is to encode the crime frame in $\hat{s}_{2,f}$, but $s_1$ and $s_3$ are from the economic frame, the model is likely to generate economic text, because it learns to reuse frame information encoded in $s_1$ and/or $s_3$ based on its experience. Inspired by adversarial learning, our third strategy is thus {\em to add ``negative'' training instances where the output sentence $\bar{s}_{2,f}$ is from the target frame}, but possible incoherent and/or topic inconsistent to the input.

In the given example, $\bar{s}_{2,f}$ would be a sentence with the crime frame. In case we combine adversarial learning with named-entity preservation, $\bar{s}_{2,f}$ is chosen from all sentences $s_2$ in a given training set, such that the named entities of $\bar{s}_{2,f}$ and $s_2$ are as similar as possible. In case not, we choose a random sentence $s_2$ as $\bar{s}_{2,f}$. Conceptually, we thereby force the model to discard any possible input frame features. We note that this learning strategy likely harms the coherence and topic consistency of the generated text, as $\bar{s}_{2,f}$ will often not fit to $s_1$ and~$s_3$. We can control this effect, though, through a careful use of the strategy, training only a few epochs.

\section{Dataset}
\label{sec:data}

In this section, we describe how we prepare the corpus we use in order to create training and test instances for the sentence-level fill-in-the-blank task.

\subsection{The Media Frames Corpus}
To analyze media framing across different social issues, \newcite{card:2015} built a corpus that comprises 35,701 news articles (published between 1990 and 2012 in 13 news portals)
in US, addressing the topics of death penalty, gun control, immigration, same-sex marriage, and tobacco.%
\footnote{We use the updated version from the authors' repository, \url{https://github.com/dallascard/media_frames_corpus}. Thus, the data distribution differs from the one of \newcite{card:2015}.}
Each article is annotated at span level for 15 general frames of the \emph{Policy Frames Codebook}~\cite{boydstun:2013} in terms of the primary frame, the title's frame, and the span-level frame. \newcite{card:2015} truncated articles to have at most 225 words.\,\,

\subsection{Data Preprocessing}

Following several works in frame analysis~\cite{naderi:2017, hartmann:2019}, we focus on the five most frequently labeled frames in the corpus, accounting for about 60\% of all labels. Examining these frames, we observed that two of them are hard to distinguish in various cases, namely {\em 6: Policy prescription and evaluation} and {\em 13: Political}.%
\footnote{\newcite{naderi:2017} reported similar observations.}
Hence, we merge those two, ending up with a set $F = \{ e, l, p, c\}$ of four frames:
\begin{itemize}
\setlength{\itemsep}{0pt}
\item[$e.$]
{\bf Economic.} Costs, benefits, or other financial implications;
\item[$l.$]
{\bf Legality, constitutionality, and jurisprudence.} Rights, freedoms, and authority of individuals, corporations, and government;
\item[$p.$]
{\bf Policy prescription and evaluation + Political.} Discussion of specific policies aimed at addressing problems, or considerations related to politics and politicians, including lobbying, elections, and attempts to sway voters;
\item[$c.$]
{\bf Crime and punishment.} Effectiveness and implications of laws and their enforcement.
\end{itemize}

For the sentence-level fill-in-the-blank task, we split the corpus articles into a training, a validation, and a test set. Each of the latter two comprises 3000 pseudo-randomly selected articles, 600 for each of the five given topics. The training set includes the remaining 29,701 articles. For each set, we collected all sentences from the respective articles that are labeled with one of the four considered frames. A sentence is considered to be labeled, if any part of the sentence is labeled. In case a sentence has more than one frame label, the sentence is associated with all the labels. For each of these framed sentences, $s_2$, we obtain its predecessor, $s_1$, and its successor $s_3$. Together, they form one data instance, as in Section~\ref{sec:approach}, where the input is the tuple of $\langle s_1, s_3 \rangle$ and the output is $s_2$. 

\begin{table}[t]

\small
\centering
\setlength{\tabcolsep}{5pt}
\begin{tabular*}{\linewidth}{ll@{}rrr}
\toprule
\bf \# & \bf Frame & \bf Training & \bf Validation & \bf Test\\
\midrule
$e$ & Economic &  6\,605 &    883 &    888 \\
$l$ & Legality c.a.j.   & 15\,313 & 1\,568 & 1\,656 \\
$p$ & Policy p.a.e. + Political & 20\,903 & 2\,169 & 2\,109 \\
$c$ & Crime    & 10\,726 & 1\,144 & 1\,257 \\
\midrule
& All four frames    &  53\,547 & 5\,764 & 5\,910 \\
\bottomrule
\end{tabular*}
\caption{The number of  fill-in-the-blank instances in the training, validation, and test set for each frame. Note that the four frames are not evenly distributed.}
\label{table-dataset}
\end{table}

To avoid that outliers mislead the learning process, we actually do not take all instances, but we filter instances by sentence length as follows. We consider only sentences $s_1$, $s_2$ and $s_3$ with at least five and at most 50 tokens each, and include only instances where $s_2$ has a similar length to the mean length of $s_1$ and $s_3$, with a tolerance of $\pm$ 50\%. About 62\% of the instances remain after this step.

The distribution of the framed sentences among the training, validation, and test sets is shown in Table~\ref{table-dataset}. Note that the test set here is the one built for the automatic evaluation. The test set for the manual evaluation is discussed in Section~\ref{subsection:m_evaluation}.

\section{Experiments}
\label{sec:experiments}

This section reports on our experiments with our reframing approach (Section~\ref{sec:approach}) on the data from Section~\ref{sec:data}. We present the results of the pilot study for the different reframing approaches, the metrics for automatic evaluation, and the design of crowdsourcing task for manual evaluation.

\subsection{Operationalizing Reframing}

We rely on transformers~\cite{wolf:2020} as the basis for reframing. The pretrained weights of the sequence-to-sequence model are from {\em T5-base}~\cite{raffel:2020}. The three strategies from Section~\ref{sec:approach} require pretraining on framed language ($\strategy_{\rm F}$) or a fine-tuning of the reframing model ($\strategy_N$ and $\strategy_{\rm A}$) respectively. For $\strategy_{\rm F}$ and $\strategy_{\rm N}$, the models were optimized on the validation set; for the adversarial learning strategy, $\strategy_{\rm A}$, we trained for three epochs in order not to harm the coherence of the output  too much. Since each strategy can be applied independently, we considered eight reframing model variations, ranging from applying no strategy ($\strategy_{\emptyset}$) to applying all three strategies ($\strategy_{\rm FNA}$).

\paragraph{Baselines}

The variant without any strategy, $\strategy_{\emptyset}$, can be considered as a baseline. Few other models exist so far that are suitable baselines for tackling the reframing task, but one is {\em GPT-2}~\cite{radford:2019}. Specifically, we finetuned GPT-2 on all text available for each frame to have four framed versions of GPT-2. During application, we used~$s_1$, the sentence before the target sentence, as the prompt and generated $s_{2, f}$ with the finetuned GPT-2. We also tested framed-language pretraining, $\strategy_{\rm F}$, with GPT-2. To obtain {\em GPT-2 + $\strategy_{\rm F}$}, we first finetuned GPT-2 on all framed text and then further finetuned it on the text of the respective frame.

\subsection{Pilot Study}

In our manual evaluation below, we focus on three of the eight variations of our approach, for budget reasons and for keeping the evaluation manageable:
\begin{enumerate}
\setlength{\itemsep}{0pt}
\item
{\em B.Coherence}. The model variation generating the most coherent sentences.
\item
{\em B.Framing}. The model variation generating the most accurately framed sentences. 
\item
{\em B.Balance}. The model variation achieving the best balance between coherence and framing. 
\end{enumerate}

We ranked the models in a pilot study where we randomly selected 10~instances $\langle s_1, s_2, s_3\rangle$ from the test set for each of the four frames in~$F$, 40~instances in total. We used the respective variation to reframe all sentences $s_2$ to the economic frame. Then, two authors of this paper were asked to judge each reframed sentence by assigning scores in response to the following questions:
\begin{enumerate}
\setlength{\itemsep}{0pt}
\item[Q1.]
Is the sentence coherent to other sentences? \\ \{\emph{yes (2) | partially (1) | no (0)}\}
\item[Q2.]
Does the sentence cover economic aspects? \\ \{\emph{yes (2) | partially (1) | no (0)}\}
\end{enumerate}

Table~\ref{table-pilot} shows the averaged scores. The Pearson's correlation $r$ for the two questions was 0.90 and 0.66 respectively, suggesting that the judges agreed substantially in the rankings. Based on the average scores, we made the following choices:

\begin{enumerate}
\setlength{\itemsep}{0pt}
\item
{\em B.Coherence}. $\strategy_{\rm FN}$ (coherence score 1.35)
\item
{\em B.Framing}. $\strategy_{\rm A}$ (framing score 0.89)
\item
{\em B.Balance}. $\strategy_{\rm NA}$ (harmonic mean 0.93)
\end{enumerate}

We chose $\strategy_{\rm NA}$ in the latter case, since it showed the maximum harmonic mean of the two scores. In addition, we manually evaluated $\strategy_{\emptyset}$, the baseline model without any training strategies.

\begin{table}[t]
\small
\centering
\setlength{\tabcolsep}{4.5pt}
\begin{tabular*}{\linewidth}{lrrrrrrr}
\toprule
						& \multicolumn{3}{c}{\bf Q1 (Coherence)}	& \multicolumn{3}{c}{\bf Q2 (Framing)}	& \multicolumn{1}{c}{\bf Balance}		\\
						\cmidrule(l@{2pt}r@{2pt}){2-4}			\cmidrule(l@{2pt}r@{2pt}){5-7}			\cmidrule(l@{2pt}r@{2pt}){8-8}
\bf Strategy 				& \bf A1	& \bf A2	& \bf Avg. 			& \bf A1	& \bf A2	& \bf Avg.			& \bf H.\ Mean		\\
\midrule
\bf $\strategy_{\emptyset}$ &4		& 6		& 0.96			& 5		& 7		& 0.49			& 0.65		\\
\addlinespace
\bf $\strategy_{\rm F}$		& 1		& 2		& 1.30			& 6		& 6		& 0.50			& 0.72		\\
 $\strategy_{\rm N}$ 		& 3		& 3		& 1.10			& 4		& 5		& 0.58			& 0.76 		\\
 $\strategy_{\rm A}$ 		& 7		& 7		& 0.50			& 1		& 2		& \bf 0.89		& 0.64		\\
\addlinespace
 $\strategy_{\rm FN}$ 		& 2		& 1		& \bf 1.35		& 7		& 2		& 0.57			& 0.80		\\
 $\strategy_{\rm FA}$ 		& 8		& 8		& 0.16			& 8		& 8		& 0.27			& 0.20		\\
 $\strategy_{\rm NA}$ 		& 5		& 4		& 0.99			& 2		& 1		& 0.88			& \bf 0.93		\\
\addlinespace
 $\strategy_{\rm FNA}$ 		& 6		& 5		& 0.90			& 3		& 2		& 0.70			& 0.79		\\
\bottomrule
\end{tabular*}
\caption{The pilot study rankings by the two annotators (A1, A2) along with the average of their scores from the eight model variations, resulting from the three training strategies $\strategy_F$, $\strategy_N$, and $\strategy_A$. Three framing variations are ranked second for A2 due to identical average scores. The right-most column shows the harmonic mean of the two average scores of both questions.}
\label{table-pilot}
\end{table}


\subsection{Evaluation Metrics}
\label{subsection:m_evaluation}

To answer the research questions, we considered three dimensions for the different approaches: coherence, correct framing, and topic consistency, both in automatic and in manual evaluation.

\paragraph{Automatic Evaluation} 

We used ROUGE scores to approximate the overall quality of the generated texts. As ROUGE requires ground-truth information, we considered only those cases where the target frame matches the frame where the test instance stems from. To quantify the effect of reframing, we compiled a vocabulary for each frame by taking the 100 words with the highest TF-IDF values, where each sentence of a frame was seen as one document. By counting the number of words occurring in the respective vocabulary, we could get a rough idea about the reframing impact.

\paragraph{Manual Evaluation} 

For the manual evaluation, we randomly selected 15~instances for each frame from the test set, 60~instances in a total. For each instance, we applied the reframing models along with baselines to reframe it to the four frames in~$F$. Among the reframed cases one was of type {\em intra-frame generation} (i.e., it had the frame from the original sentence); the other cases were of the {\em inter-frame generation} type. These two types will be discussed separately.

We used Amazon Mechanical Turk to evaluate the selected test set, where each instance was annotated by five workers (for \$0.80 per instance). For reliability, we employed only master workers with more than 95\% approval rate and more than 10k approved HITs. The percentage of the agreement to the majority is 73\% on average in our experiments.  The workers were provided three continuous sentences and were asked to judge the middle one (the one generated) by answering six  questions:
\begin{enumerate}
\setlength{\itemsep}{0pt}
\item[Q1.]
Is the sentence coherent to other sentences? \\
\{\emph{yes (2) | partially (1) | no (0)}\}
\item[Q2.]
Does the sentence match the topic in the first and the last sentence?  \\
\{\emph{Same or close related topic (2) | related or no topic (1) | unrelated topic (0)}\}
\item[Q3.]
Does the sentence cover economic aspects? \\
\{\emph{yes (2) | partially (1) | no (0)}\}
\item[Q4.]
Does the sentence cover legality-related aspects? \\
\{\emph{yes (2) | partially (1) | no (0)}\}
\item[Q5.]
Does the sentence cover policy-related aspects? \\
\{\emph{yes (2) | partially (1) | no (0)}\}
\item[Q6.]
Does the sentence cover crime-related aspects? \\
\{\emph{yes (2) | partially (1) | no (0)}\}
\end{enumerate}

The first two questions asked for coherence and topic consistency, respectively. The latter four assessed the reframing effect. For the computation of the framing scores presented below, only the question asking for the target frame was taken into account. Since a sentence may serve multiple frames, the four framing questions were asked individually. We believe this scoring method is better than only asking whether a text has a desired frame, to avoid making the question suggestive.
Along with this questionnaire, the definition of the four frames were provided.

\section{Results and Discussion}
\label{sec:discussion}

This section discusses the automatic and manual evaluation results, in order to then analyze how our three training strategies affect generation. Finally, we show some examples from the reframed output and discuss the limitations of our approach.

\begin{table}[t]
\small
\centering
\setlength{\tabcolsep}{4.5pt}
\begin{tabular*}{\linewidth}{l@{}rrrrr}
\toprule
			& \multicolumn{3}{c}{\bf (a) w/ Entities}		& \multicolumn{2}{c}{\bf (b) w/o Entities} \\
			\cmidrule(l@{2pt}r@{2pt}){2-4}			\cmidrule(l@{2pt}r@{2pt}){5-6}	
\bf Approach & \bf Rou.-1 & \bf Rou.-2 & \bf Rou.-L & \bf Rou.-1 & \bf Rou.-L  \\
\midrule
$\strategy_{\emptyset}$                                   & 16.37 &  2.90 & 13.48 & 13.51 & 11.22 \\
\addlinespace
$\strategy_{\rm F}$                                             & 16.02 &  2.61 & 13.00 & 14.37 & 11.84 \\
$\strategy_{\rm N}$                                             & 27.06 & 10.44 & 23.83 & 14.91 & 12.62 \\
$\strategy_{\rm A}$                                             &  9.78 &  0.61 &  8.13 &  9.68 &  8.20 \\
\addlinespace
$\strategy_{\rm FN}$                       & \bf 29.70 & \bf 12.32 & \bf 26.27 & \bf 16.42 & \bf 13.97 \\
$\strategy_{\rm FA}$                       & 11.47 &  0.62 &  9.30 & 11.48 &  9.38 \\
$\strategy_{\rm NA}$                       & 24.54 &  9.25 & 21.72 & 12.04 & 10.22 \\
\addlinespace
$\strategy_{\rm FNA}$ & 25.83 & 10.36 & 23.01 & 12.58 & 10.64 \\
\midrule
GPT-2       & 11.97 & 1.14 & 9.80 & 10.66 & 8.96 \\ 
GPT-2\,+\,$\strategy_{\rm F}$ & 12.06 & 1.16 & 9.85 & 10.74 & 9.00 \\ 
\bottomrule
\end{tabular*}
\caption{Rouge-1, Rouge-2, and Rouge-L $F_1$-scores (a)~with and (b)~without considering named entities of all model variations (based on our strategies $\strategy_{\rm F}$, $\strategy_{\rm N}$, and $\strategy_{\rm A}$) compared to the GPT-2 baselines. Rouge-2 is ignored for (b), since entity removal makes it unreliable. The highest score in each column is marked bold.}
\label{table-rouge}
\end{table}

\begin{table}[t]

\small
\centering
\setlength{\tabcolsep}{7pt}
\begin{tabular*}{\linewidth}{llll}
\toprule
\bf Economic ($e$) & \bf Legality ($l$) & \bf Policy ($p$) &  \bf Crime ($c$) \\
\midrule
tobacco & court & gun & death \\
said & said & said & said \\
gun & state & bill & gun \\
would & marriage & would & police \\
state & death & state & murder \\
million & law & marriage & year \\
new & sex & law & penalty \\
industry & supreme & house & law \\
year & judge & ban & state \\
smoking & same & new & two \\
\bottomrule
\end{tabular*}
\caption{The top-10 words having the highest TF-IDF values for each of the four frame in $F = \{ e, l, p, c\}$.}
\label{table-framing-words}
\end{table}

\begin{table}[t]

\small
\centering
\setlength{\tabcolsep}{3pt}
\begin{tabular*}{\linewidth}{lllll}
\toprule
\bf Approach & \bf Economic & \bf Legality & \bf Policy & \bf Crime   \\
\midrule
$\strategy_\emptyset$ & 10\% ($-$2) & 12\% ($-$1) & 12\% ($-$1) & 11\% ($-$2) \\ 
\addlinespace          
$\strategy_{\rm F}$   & 11\% ($-$1) & 13\% ($+$0) & 12\% ($+$0) & 11\% ($+$0) \\
$\strategy_{\rm N}$   & 11\% ($-$1) & 13\% ($+$0) & 12\% ($+$0) & 12\% ($-$1) \\
$\strategy_{\rm A}$   & 15\% ($+$2) & 20\% ($+$6) & 12\% ($+$0) & 15\% ($+$1) \\
\addlinespace
$\strategy_{\rm FN}$  & 11\% ($-$1) & 13\% ($+$0) & 12\% ($+$0) & 12\% ($-$1) \\
$\strategy_{\rm FA}$  & 17\% ($+$4) & 17\% ($+$3) & 18\% ($+$5) & 13\% ($+$0) \\
$\strategy_{\rm NA}$  & 13\% ($+$0) & 18\% ($+$4) & 16\% ($+$3) & 15\% ($+$2) \\
\addlinespace
$\strategy_{\rm FNA}$ & 12\% ($+$0) & 19\% ($+$5) & 16\% ($+$3) & 17\% ($+$4) \\
\midrule
GPT-2                        & \phantom{0}8\% ($-$4) & 10\% ($-$3) & 10\% ($-$2) & \phantom{0}9\% ($-$3) \\ 
GPT-2 + $\strategy_{\rm F}$  & \phantom{0}9\% ($-$3) & 10\% ($-$3) & 10\% ($-$2) & \phantom{0}9\% ($-$3) \\  
\bottomrule
\end{tabular*}
\caption{Proportion of word overlaps between the reframed texts and the top-100 TF-IDF words of all four frames for each model variation and the GPT-2 baselines. The numbers in parentheses show the difference to the texts before reframing (in percentage points).}
\label{table-framing-words-overlaps}
\end{table}

\begin{table}[t]

\small
\centering
\setlength{\tabcolsep}{2pt}
\begin{tabular*}{\linewidth}{@{}l rrrr rrrr@{}}
\toprule
& \multicolumn{4}{c}{\bf Intra-Frame} &\multicolumn{4}{c}{\bf Inter-Frame}\\
\cmidrule(r@{\tabcolsep}l@{\tabcolsep}){2-5} \cmidrule(r@{\tabcolsep}l@{\tabcolsep}){6-9}
& topic & coh. & fram. & avg & topic & coh. & fram. & avg \\
\midrule
B.Coherence                 & 1.63     & \bf 1.71 & 1.59     & \bf 1.64 & \bf 1.64 & \bf 1.68 & 1.60 & \bf 1.64\\
B.Framing                   & 1.59     & 1.65     & \bf 1.65 & 1.63     & 1.58     & 1.61 & \bf 1.64 & 1.61\\
B.Balance                   & 1.57     & 1.61     & 1.62     & 1.60     & 1.56     & 1.63 & 1.62 & 1.60\\
\midrule
GPT-2 + $\strategy_{\rm F}$ & 1.54     & 1.61     & 1.57     & 1.57     & 1.55     & 1.59 & 1.58 & 1.57\\
$\strategy_{\emptyset}$                        & \bf 1.66 & 1.66     & 1.61     & 1.64     & 1.63     & 1.66 & 1.60 & 1.63\\
\bottomrule
\end{tabular*}
\caption{Manual evaluation: The {\em topic} consistency, {\em coh}erence, {\em fram}ing, and average scores ({\em avg}) in intra- and inter-frame generation for the model varations with highest coherence ($\strategy_{\rm FN}$), framing ($\strategy_{\rm A}$), and balanced ($\strategy_{\rm NA}$) scores in the pilot study, compared to baselines. The best score in each column is marked bold. }
\label{table-crowdsourcing}
\end{table}

\begin{table}[t]
\small
\centering
\setlength{\tabcolsep}{2pt}
\begin{tabular*}{\linewidth}{lrrrrr@{\quad}rrrrr}
\toprule
&\multicolumn{5}{c}{\bf Coherence of $s_{2,f}$} &\multicolumn{5}{c}{\bf Framing of $s_{2,f}$}\\
\cmidrule(r@{\tabcolsep}l@{\tabcolsep}){2-6} \cmidrule(r@{\tabcolsep}l@{\tabcolsep}){7-11}
$s_2$ 	& $e$ & $l$ & $p$ & $c$ & \bf avg & $e$ & $l$ & $p$ & $c$ & \bf avg\\
\midrule
$e$ & --         & 1.71  & \bf 1.79  & 1.69 & 1.73 & --        & 1.65     & 1.59 & 1.59 & 1.61\\
$l$ & 1.71      & --     & 1.63      & 1.67 & 1.67 & \bf 1.75 & --        & 1.55 & 1.56 & 1.62\\
$p$ & 1.67      & 1.68  & --         & 1.68 & 1.68 & 1.63     & 1.68     & --    & 1.61 & 1.64\\
$c$ & \it 1.57  & 1.67  & 1.65       & --   & 1.63 & 1.53     & \it 1.48 & 1.56 & --    & 1.52\\
\midrule
\bf avg & 1.65    & 1.69  & 1.69     & 1.68 & 1.68 & 1.64     & 1.60     & 1.57 & 1.59  & 1.60  \\
\bottomrule
\end{tabular*}
\caption{Manual evaluation: The average coherence and framing scores of reframing from $s_2$ to $s_{2, f}$ for each pair of source fram (rows) and target frame (columns) from $\{e, l, p, c\}$. The highest/lowest score of each dimension is marked bold/italic.}
\label{table-directions}
\end{table}

\begin{table}[t!]
	
\small
\centering
\setlength{\tabcolsep}{2pt}

\begin{tabular}{p{0.98\linewidth}}
\toprule
{Original text} \\
\midrule
All that was true of interracial marriages shortly after World War II. 
\textbf{Today, interracial marriage has strong public support, and no successful politician or prominent public figure favors outlawing such unions.} 
The question is whether gay marriage is on the same trajectory or is so fundamentally different that it will never be legalized.\\
\addlinespace
\midrule
{\bfseries (b)~GPT-2 + $\strategy_{\rm F}$} (reframed text) \\
\midrule
All that was true of interracial marriages shortly after World War II. 
\textbf{Actually, a number of people within a decade may have wanted to apply for visas. The ''Hispanic} 
The question is whether gay marriage is on the same trajectory or is so fundamentally different that it will never be legalized.\\
\addlinespace
\midrule
{\bfseries (c)~$\strategy_{\emptyset}$} (reframed text) \\
\midrule
All that was true of interracial marriages shortly after World War II. 
\textbf{Bob Greene: Gay marriage is on the same trajectory as interracial marriage. He says it is so fundamentally different that it will never be legalized.} 
The question is whether gay marriage is on the same trajectory or is so fundamentally different that it will never be legalized.\\
\addlinespace
\midrule
{\bfseries (d)~B.Coherence} (reframed text) \\
\midrule
All that was true of interracial marriages shortly after World War II. 
\textbf{Today, same-sex marriages are legal in many states, but they are not as legal as interracial marriages, according to a new study.} 
The question is whether gay marriage is on the same trajectory or is so fundamentally different that it will never be legalized.\\
\addlinespace
\midrule
{\bfseries (e)~B.Framing} (reframed text) \\
\midrule
All that was true of interracial marriages shortly after World War II. 
\textbf{``It's a good thing that we're able to do this,'' said a spokesman for the tobacco industry, who is suing the tobacco companies.} 
The question is whether gay marriage is on the same trajectory or is so fundamentally different that it will never be legalized.\\
\addlinespace
\midrule
{\bfseries (f)~B.Balance} (reframed text) \\
\midrule
All that was true of interracial marriages shortly after World War II. 
\textbf{Today's tight labor market dictates that employers consider workers based on the skills they possess rather than the partners they prefer. Gay couples must also consider the financial obligations they owe their employers, he says.} 
The question is whether gay marriage is on the same trajectory or is so fundamentally different that it will never be legalized.\\
\bottomrule
\end{tabular}

\caption{
(a)~Sample text from the media frames corpus \cite{card:2015}. The bold sentence is labeled with the {\em policy} frame.
(b-f)~Reframed sentences with the five manual labeled approaches to the {\em economic} frame.}
\label{table-reframing-example2}

\end{table}

\subsection{Automatic Evaluation}

We here use ROUGE to assess the similarity between the generated text and the ground-truth text. As some model variations use named entities extracted from the ground truth, we also consider a ROUGE variation where named-entity matches are ignored in the computation.

Table~\ref{table-rouge} shows the results. We see that the GPT-2 baselines perform worse than most model variations in all ROUGE scores. Adding the \emph{framed-language pretraining} strategy ($\strategy_{\rm F}$) improves GPT-2 to some extent, though. The other two strategies cannot be applied directly to GPT-2. When using either strategy in isolation, only \emph{named-entity preservation ($\strategy_{\rm N}$)} improves the ROUGE scores over $\strategy_\emptyset$. Even though $\strategy_{\rm N}$ learns to reuse the named entities from the ground-truth texts, we also see some improvement for ROUGE without named entity overlaps. Using only \emph{adversarial learning} ($\strategy_{\rm A}$) decreases the ROUGE scores the most. This matches our expectation that $\strategy_{\rm A}$ harms coherence. 

Among the strategy combinations, $\strategy_{\rm FN}$ has the highest ROUGE score both with and without named entity overlaps. This suggests that $\strategy_{\rm F}$ and $\strategy_{\rm N}$ are important to generate texts of good quality. By contrast, $\strategy_{\rm A}$ tends to decrease the ROUGE scores also here, for example, comparing $\strategy_{\rm F}$ with $\strategy_{\rm FA}$. 
Note, however, that ROUGE tells us little about the correct framing.

\paragraph{Framing Word Overlaps}

Table~\ref{table-framing-words} lists the top-10 framing words in each frame. Some words are characteristic for more than one frame, such as ``gun'' ({\em Economic} and {\em Crime}). Via manual inspection, we found that the economic frame covers the gun-sailing market while the crime frame tackles gun-control issues. The frames also have distinctive words, such as ``industry'' ({\em Economic}), ``judge'' ({\em Legality}), ``bill'' ({\em Policy}), and ``police'' ({\em Crime}). 

Table~\ref{table-framing-words-overlaps} shows the proportions of framing words used in the test set, before and after reframing. It becomes clear that the variations including {\em adversarial learning} ($\strategy_{\rm A}$) increase the number of framing words the most. GPT-2 models generated even fewer framing words in each frame.

\subsection{Manual Evaluation}

\paragraph{Intra-Frame Generation}

We first look at those generated sentences $s_{2,f}$ where the target frame~$f$ is the frame used in the ground-truth, $s_2$. Intra-frame generation can be seen as easier for a reframing model, since some frame information may be leaked in the previous or the next sentences. 

The left block of Table~\ref{table-crowdsourcing} shows the results. GPT-2 + $\strategy_{\rm F}$ is worst in almost every case. In terms of keeping the topic consistent, the best approach is $\strategy_\emptyset$. For coherence scores, however, {\em B.Coherence} ($\strategy_{\rm FN}$) obtains the highest averaged coherence score (1.71), as expected from the pilot study. Similarly, the best one for framing (1.65) is {\em B.Framing} ($\strategy_{\rm A}$). The high consistency between the pilot study judges and the crowdsourcing workers speaks for the reliability of the results. With an average score of 1.64, {\em B.Coherence}, is, with tiny margin, the best among all approaches in intra-frame generation.

\paragraph{Inter-Frame Generation} 

Inter-frame generation requires an actual {\em re}framing. Its results are shown in the right block of Table~\ref{table-crowdsourcing}. Similar to intra-frame generation, the most coherent sentences were generated by {\em B.Coherence} (1.68), which is also best for topic consistency (1.64) this time, slightly outperforming $\strategy_\emptyset$. Overall, the best model in the inter-frame generation is {\em B.Coherence} again. {\em B.Balance} ($\strategy_{\rm FN}$) is the third-best in coherence and the second-best in framing, but due to its comparably low topic-consistency score (1.56), it is the worst variation on average.

Taken together, the tiny but important difference between the intra- and inter-frame generations lies in the fact that $\strategy_\emptyset$ performs better in the intra-frame generation than in the other. This suggests that, while the baselines are useful in easier cases, in the actual reframing task our proposed strategies are still needed. Besides, we observe that the inter-frame generation scores are just slightly lower than those in intra-frame generation. Considering that reframing is notably more complicated than generating the same frame, we conclude that our model realizes our reframing goals well in principle. Altogether, the rather high scores suggest that the neural generation models perform strong in general---or that our crowdworkers were not critical enough.

To get further insights in Table~\ref{table-directions}, we take a closer look at the different reframing directions (source frame to target frame), focusing on the best overall model in Table~\ref{table-crowdsourcing}, {\em B.Coherence}. We find that it seems rather difficult to change crime-framed sentences (source $c$) to other frames, especially changing it to {\em Economic} ($e$). This observation may be explained by the low word overlap between {\em Crime} and other frames. On the contrary, changing the {\em Policy} frame ($p$) to others seems to work better on average. When discussing policies in context, it may be easier for models to add side effects regarding economics or crime, while this is not the case for other source frames.

\subsection{Training Strategies}
\label{subsection:discussion_training_strategies}

\paragraph{Framed-Language Pretraining ($\strategy_{\rm F}$)} 

Comparing GPT-2 and GPT-2 + $\strategy_{\rm F}$ in Table~\ref{table-rouge}, we observe that using $\strategy_{\rm F}$ can slightly improve the text quality in terms of ROUGE scores. However, the benefits of this training strategy are more obvious when combining it with $\strategy_{\rm N}$. For example, $\strategy_{\rm FN}$ has about two percentage ROUGE higher compared to $\strategy_{\rm F}$.

\paragraph{Named-Entity Preservation ($\strategy_{\rm N}$)} 

To generate a coherent and topic-consistent text, preserving named entities turns out to be very important. In terms of ROUGE, strategy $\strategy_{\rm N}$ is the most powerful feature. On the other hand, the model achieving the highest topic and coherence score according to the crowdsourcing results in Table~\ref{table-crowdsourcing}  (B.Coherence) also uses this strategy, together with $\strategy_{\rm F}$.

\paragraph{Adversarial Learning ($\strategy_{\rm A}$)} 

In terms of neither automatic nor manual evaluation, applying adversarial learning gives any improvement to the text quality. However, including it can generate better framed text: In both the pilot study and the crowdsourcing study, including adversarial learning resulted in the highest framing scores.

\subsection{Examples}

Table~\ref{table-reframing-example2} exemplifies the effect of sentence-level reframing, showing how the five manually evaluated models reframed a text from the policy to the economic frame. In this particular example, the intituively little connection between the topic of gay marriage and the frame of economy makes the reframing task particularly challenging. 

As the table shows, only two models successfully managed to change the focus, {\em B.Framing} and {\em B.Balance}. In particular, the result of the former mentions an opinion of ``a spokesman for the tobacco industry'', the latter uses the labor market's viewpoint. However, the text ``It's a good thing that we're able to do this'' in {\em B.Framing} appears to be rather vague and general. Besides, the text is related to economy only because it mentions the tobacco industry. On the other hand, {\em B.Balance} integrates gay marriage and economy in a more natural way by using the labor market to connect the two concepts.

\section{Conclusion}
\label{conclusion}

Unlike several existing studies, where style transfer is addressed at the word or phrase level, this paper studies {\em sentence-level} style transfer for the problem of reframing news articles. We have cast this problem as a sentence-level fill-in-the-blank task, generating new sentences with target frames while maintaining their coherence and topic consistency with the surrounding context. To tackle the task, we have proposed three training strategies to control the framing and coherence of the generated sentences. Evaluating these strategies automatically and manually, we found that, although it is not possible for any single strategy to fulfill the needs of reframing, combining the strategies leads to a successful reframing of news articles with reasonable coherence and topic consistency.

Even though we are aware of the limitation of our approach, we argue that such a sentence-level reframing is a big step towards full article reframing. In future work, we plan to cover a more fine-grained set of frames, and to consider reframing at the level of paragraphs or entire articles.

\section{Ethical Concerns}
\label{subsection:limitations}

We are aware of the ethical concerns raised by our approach. Especially, generated sentence parts such as quotes may not be correct factually. Exemplary, we generated the sentence {``It's a good idea,'' said Sen. John McCain, D-N.Y.}, whereas John McCain was neither a Democrat nor a Senator from New York. Also, it is unlikely that John McCain had said that in the given context. In a real-world scenario, users would have to validate the truth of the generated texts, which we cannot expect from them in general. Given that we do not see our model as mature enough for application yet, we leave more elaborated solutions to this problem to future work, but we clearly point out that computationally reframed sentences should be marked as such to people working with respective technology.

\section*{Acknowledgments}

This work was partially supported by the German Research Foundation (DFG) within the Collaborative Research Center ``On-The-Fly Computing'' (SFB~901/3) under the project number~160364472.

\bibliography{emnlp21-reframing-lit}
\bibliographystyle{acl_natbib}

\end{document}